\documentclass[5p,times]{elsarticle}

\usepackage{amssymb}
\usepackage{amsmath}
\usepackage{graphicx}
\usepackage{booktabs,tabularx,array}
\usepackage{xcolor}
\usepackage{xurl}
\usepackage{stfloats}
\usepackage{microtype}
\usepackage{balance}

\usepackage[bookmarks=false]{hyperref}
    \hypersetup{colorlinks,
      linkcolor=blue,
      citecolor=blue,
      urlcolor=blue}
\biboptions{sort&compress}

\graphicspath{{figures/}}
\newcolumntype{Y}{>{\raggedright\arraybackslash}X}

\makeatletter
\gdef\emailauthor#1#2{\stepcounter{ead}%
  \g@addto@macro\@elseads{\raggedright\let\corref\@gobble
    \eadsep{#1} (#2)\def\eadsep{\unskip,\space}}}
\AtBeginDocument{\g@addto@macro\@elseads{\unskip.}}
\makeatother

\begin{document}

\begin{frontmatter}

\title{Closed-loop weld lobe development in resistance spot welding using large language models}

\author[a]{Zhaochen Yang}
\author[b]{Wayne Cai}
\author[b]{Venkat Aitharaju}
\author[b]{Hassan Ghassemi-Armaki}
\author[a]{Jeffrey A. Abell\corref{cor1}}
\ead{jaabell@umich.edu}
\author[a]{Chenhui Shao\corref{cor1}}
\ead{chshao@umich.edu}

\address[a]{Department of Mechanical Engineering, University of Michigan, Ann Arbor, MI 48109, USA}
\address[b]{General Motors Global Research \& Development, Warren, MI 48092, USA}

\begin{abstract}
Weld lobe development is an important task in industrial resistance spot welding (RSW) because the lobe defines the combinations of welding current and welding time that produce acceptable welds.  Welding engineers find the lobe boundaries through repeated trial welds and must repeat the trials whenever materials, equipment, or operating conditions change, which costs time, material, and labor.  Large language models (LLMs) have been applied to manufacturing tasks such as knowledge retrieval, question answering, and process data analysis, but it is not known whether they can make quantitative process decisions.  This paper proposes a closed-loop weld lobe development method for RSW using LLMs.  In each iteration, the LLM selects the welding schedule for the next trial weld and updates its lobe prediction from the trial results collected so far.  No task-specific training is required.  We evaluate the method with Gemma, Claude Sonnet 5.5, and GPT-5.6 on eleven weld lobes from four open-access studies, using a virtual trial-welding laboratory built from the source data to return trial outcomes.  After 30 trial welds, the mean interval intersection over union (IoU) between the predicted and reference lobes is 0.695, 0.902, and 0.821 for Gemma, Claude, and GPT, respectively.  Two machine-learning baselines, Bayesian information-gain selection with Bayesian estimation and proximal policy optimization with Deep Sets, reach 0.871 and 0.769.  Without trial feedback, the same LLMs recommend schedules poorly, with at most 33.3\% of the recommended schedules falling inside the lobe.  General-purpose LLMs can therefore develop weld lobes from trial feedback about as accurately as specialized machine-learning methods.  The same approach could be used for other process development tasks in manufacturing where a process window is found through sequential experiments.
\end{abstract}


\cortext[cor1]{Corresponding authors.}

\end{frontmatter}
\thispagestyle{plain}

\section{Introduction}
\label{sec:introduction}

Resistance spot welding (RSW) joins overlapping metal sheets by resistance heating under electrode force, and the welding schedule sets the process parameters~\citep{tang2003influence,morales2021}.  For a given set of materials and electrodes, the weld lobe is the region of welding current and welding time that produces acceptable welds.  Below its lower boundary the joint does not meet the required performance, and above its upper boundary expulsion occurs~\citep{zhang2000statistical,shin2019}.  Production schedules are chosen from inside the lobe, so weld lobe development is an important task in industrial RSW.

Welding engineers find the lobe boundaries through repeated trial welds~\citep{li2001statistical,cho2006design}.  A change in materials, operating conditions, or welding equipment calls for more trials, and each round costs time, materials, and labor.  A method that selects the trial welds and estimates the lobe automatically would reduce this effort.

Large language models (LLMs) have been used in manufacturing to retrieve technical information, answer questions, and organize knowledge.  \citet{freire2024}, for example, evaluate an assistant built around factory documents and issue reports.  LLMs can also turn natural-language requests into machine instructions and interpret process observations for engineers.  In additive manufacturing, they have been benchmarked on domain-specific tasks~\citep{eslaminia2025fdm} and used as agents to detect anomalies before printing~\citep{eslaminia2026llm}.  For process optimization, \citet{samani2026} combine LLM-generated guidance with Bayesian optimization in fused deposition modeling.  Process decisions, however, depend on knowledge of specific materials, equipment, and operating conditions that LLMs may lack.  It is not known whether an LLM can carry out quantitative process development on its own.

This paper proposes a closed-loop weld lobe development method for RSW using LLMs.  Unlike approaches that pair an LLM with a conventional algorithm, the LLM here runs the whole loop.  In each iteration, it selects the welding schedule for the next trial weld, receives the weld outcome, and updates its lobe prediction from all trial results so far.  We call this iterative process lobe exploration.  The case description, acceptance requirement, and trial results are given to the LLM as text, and the LLM needs no task-specific training.

The method is evaluated with three LLMs: Gemma, Claude Sonnet 5.5, and GPT-5.6.  The test cases are eleven reference lobes that we construct from measurements and boundary data in four open-access studies.  The schedules an LLM proposes may not appear in these datasets, so we build a virtual trial-welding laboratory from the source data to estimate the trial outcomes.  Lobe accuracy is measured by the mean intersection over union (IoU) between the predicted and reference current intervals at selected welding times, and we track it as trials accumulate.  The LLMs are compared with machine-learning (ML) methods on both tasks in the loop.  For trial-schedule selection, the baselines are Bayesian information gain~\citep{chaloner1995} and proximal policy optimization (PPO)~\citep{ppo2017}.  For lobe estimation, they are a Bayesian lobe model and Deep Sets~\citep{deepsets2017}.  We also evaluate schedule generation and outcome prediction without target-condition trial results, schedule revision after one failed trial, lobe transfer across electrode-force conditions, and image-based weld inspection.  Finally, we test whether retrieval-augmented generation (RAG) improves performance.

After 30 trial welds without RAG, the mean IoU is 0.695, 0.902, and 0.821 for Gemma, Claude, and GPT, respectively.  The two ML closed loops reach 0.871 (information-gain selection with Bayesian estimation) and 0.769 (PPO with Deep Sets).  Without trial feedback, at most 33.3\% of the schedules that the LLMs recommend fall inside the lobe, and RAG does not consistently improve the results.  Lobes developed at two electrode forces can also be used to predict the lobe at a third force without new trial welds.  The mean IoUs for this transfer are 0.588, 0.762, and 0.677 for the three LLMs.

These results show that general-purpose LLMs can develop weld lobes from trial feedback about as accurately as specialized ML methods.  They do so without the predefined lobe representations or synthetic training data that the ML methods rely on.  The same approach could be used for other process development tasks in manufacturing where a process window is found through sequential experiments.

This study makes three contributions:
\begin{enumerate}
\item A closed-loop weld lobe development method for RSW is proposed, in which an LLM selects the trial-weld schedules and updates the lobe prediction from the trial results without task-specific training.
\item A virtual trial-welding laboratory built from eleven published weld lobes is used to compare three LLMs with two ML baselines, and cross-evaluation separates trial selection from lobe estimation.
\item Tests of schedule generation, outcome prediction, schedule revision, lobe transfer, and image-based inspection, with and without RAG, show where LLMs are and are not reliable in RSW process decisions.
\end{enumerate}

The remainder of this paper is organized as follows.
Section~\ref{sec:methodology} presents the methodology, Section~\ref{sec:results} reports the results, Section~\ref{sec:discussion} discusses the findings and limitations, and Section~\ref{sec:conclusion} concludes the paper.

\section{Methodology}
\label{sec:methodology}

This section presents the proposed closed-loop weld lobe development method and the evaluation of its performance.

\subsection{Closed-loop weld lobe development method}
\label{sec:method}

\begin{figure}[h]
  \centering
  \includegraphics[width=0.9\columnwidth]{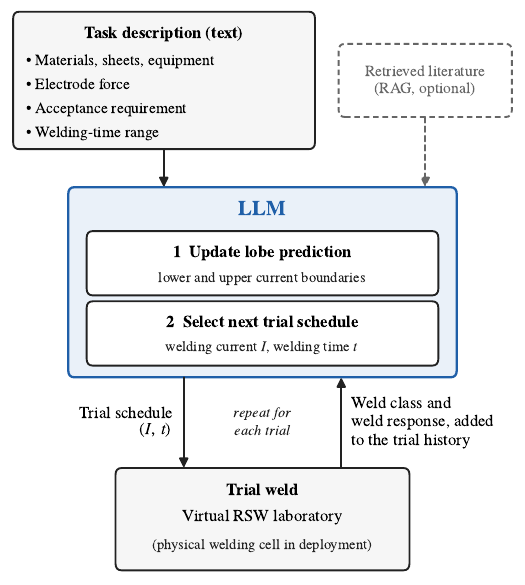}
  \caption{Proposed closed-loop weld lobe development method.  In each iteration, the LLM updates its lobe prediction and selects the next trial schedule, and the trial result is added to the trial history.}
  \label{fig:method}
\end{figure}

Figure~\ref{fig:method} illustrates the method.  The LLM receives a text description of the welding case, which gives material information, sheet thicknesses and coatings, welding equipment and electrode specifications, and process settings such as electrode force, current-pulse structure, and squeeze and hold times.  It also receives the acceptance requirement, including the minimum tensile-shear load or nugget diameter and the absence of expulsion, and the supported welding-time range.  The LLM then proposes an initial lobe and the first trial schedule.  After each trial weld, it receives the weld class and, where available, the quantitative weld response.  The LLM uses the accumulated trial records and previous lobe predictions retained in the conversation to progressively update its lobe prediction and select the next trial schedule.  This lobe exploration runs for 30 trial welds in one conversation.

The LLM reports the lower and upper current boundaries of the lobe at eleven evenly spaced welding times within the supported range, including both endpoints.

\subsection{Reference lobes and virtual RSW laboratory}
\label{sec:virtual-laboratory}

Closed-loop lobe exploration with LLMs requires feedback on the proposed welding schedules for trials. The data are taken from four open-access RSW studies~\citep{shin2019,kim2023,choi2019,morales2021}. These data contain discrete trial-weld measurements or lobe boundary points that may not exactly match the schedules proposed by the LLMs. We therefore construct a virtual RSW laboratory to estimate trial-weld outcomes, reporting the weld class (undersized, acceptable, or expulsion) and, where available, a quantitative weld response (tensile-shear load or nugget diameter).

\begin{figure}[!tbp]
  \centering
  \includegraphics[width=\columnwidth]{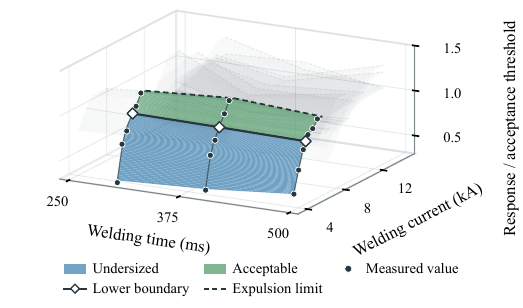}
  \caption{Reconstructed response surfaces for nine reference lobes: seven Shin cases, one Kim case, and one Choi case~\citep{shin2019,kim2023,choi2019}. The Kim case is highlighted in blue (undersized) and green (acceptable), with the other eight cases shown in gray. The surface heights represent quantitative weld responses normalized by the corresponding acceptance thresholds.}
  \label{fig:virtual-laboratory}
\end{figure}

\begin{table*}[!t]
\centering
\caption{Reference weld lobes and acceptance criteria.  $t_s$ is the sheet thickness (1.2~mm for Choi).}
\label{tab:datasets}
\footnotesize
\renewcommand{\arraystretch}{1.15}
\begin{tabularx}{\textwidth}{@{}l Y c c l Y l@{}}
\toprule
Source & Material system & Electrode force (kN) & Lobes & Weld response & Acceptance criterion & Feedback to LLM \\
\midrule
\citet{shin2019} & Aluminum--steel joints with process tapes & 1.5--4.5 (first stage) & 7 & Tensile-shear load & $\geq 3000$~N & Class and load \\
\citet{kim2023} & Conventional 316L stainless steel & 3.4 & 1 & Nugget diameter & $\geq 5.0$~mm & Class and diameter \\
\citet{choi2019} & 980~MPa-class cold-rolled steel & 2.94 & 1 & Nugget diameter & $\geq 4\sqrt{t_s}\approx4.38$~mm & Class and diameter \\
\citet{morales2021} & CP1000 steel, converter frequencies of 1 and 7~kHz & 4.7 & 2 & None (boundaries only) & Reported $I_{D\min}$ and $I_{\mathrm{Proy}}$ & Class only \\
\bottomrule
\end{tabularx}
\end{table*}

Eleven reference weld lobes are selected from the four datasets, as summarized in Table~\ref{tab:datasets}.  The Morales dataset provides only lobe boundaries, without quantitative weld-response measurements.  A weld is acceptable if it meets the criterion in Table~\ref{tab:datasets} without expulsion.  For Morales, the lobe is bounded by the reported minimum-diameter current boundary $I_{D\min}$ and expulsion current boundary $I_{\mathrm{Proy}}$.  The Shin load threshold and Choi diameter criterion are adopted for this evaluation.

For the lobes from the Shin, Kim, and Choi datasets, we estimate how the quantitative weld response varies with welding time and current to provide feedback to the LLMs. At each measured welding time, we fit a smooth, non-decreasing current--response curve using non-negative least squares with Gaussian cumulative distribution basis functions. The lowest current meeting the acceptance requirement defines the lower lobe boundary, and the first reported expulsion current defines the upper boundary where observed. Between measured welding times, we linearly interpolate the boundaries and response values. These curves and interpolated values define a surface relating welding time and current to the quantitative weld response. For each proposed trial within the supported welding-time and current ranges, the laboratory determines the weld class from the reconstructed lobe boundaries and, where available, obtains the quantitative response from the surface. These results are returned to the LLM as trial-weld feedback. For Morales, only class feedback is provided. Figure~\ref{fig:virtual-laboratory} shows the reconstructed surfaces.

\subsection{Models, RAG configuration, and evaluation metric}
\label{sec:information-protocol}

Three LLMs are evaluated: Gemma (12B), Claude Sonnet 5.5, and GPT-5.6. Gemma runs locally through Ollama, while Claude and GPT are accessed through APIs.

We compare LLMs without RAG (LLM) and with RAG (LLM+RAG). Both configurations receive the same task description as the prompt. LLM+RAG adds passages retrieved from a corpus of twelve open-access RSW papers, which cover quality assurance and monitoring~\citep{stavropoulos2024,martin2022}, effects of process parameters on weld formation and quality~\citep{song2023,ullrich2024,taufiqurrahman2021,rajarajan2022}, welding schedule design and process control~\citep{bamberg2021,bamberg2026,soomro2021,lee2020,yu2018}, and electrical resistance~\citep{fezer2025}.  The four studies used to build the reference lobes are not part of this corpus.

Predicted lobes are evaluated at the welding times reported in the source studies. Let $W_j$ and $\widehat{W}_j$ denote the reference and predicted current intervals at evaluation time $t_j$. The mean interval intersection over union (IoU) is
\begin{equation}
  Q = \frac{1}{m}\sum_{j=1}^{m}
  \frac{|W_j \cap \widehat{W}_j|}
       {|W_j \cup \widehat{W}_j|},
  \label{eq:window-iou}
\end{equation}
where $m$ is the number of evaluated welding times and $|\cdot|$ denotes interval length. This metric is used to evaluate lobe predictions in exploration and transfer.  In lobe exploration, the IoU is tracked as trials accumulate, and the prompts do not identify the welding times used for scoring.  Both LLM and LLM+RAG configurations are evaluated on all eleven lobes.

\subsection{Comparison with ML methods}
\label{sec:sequential-method}

We compare the LLM and LLM+RAG loops with two ML closed loops. The Bayesian method uses information gain for trial selection and posterior median boundaries for lobe estimation. The second uses proximal policy optimization (PPO)~\citep{ppo2017} for trial selection and Deep Sets~\citep{deepsets2017} for lobe estimation. Training and hyperparameter tuning use synthetic lobes only. PPO rewards are based on improvements in the Deep Sets lobe IoU. All lobe predictions are evaluated using mean interval IoU at the welding times reported in the source studies.

\begin{table}[!tbp]
\centering
\caption{Trial selection $\rightarrow$ lobe estimation. IG denotes information-gain selection, and Bayes denotes Bayesian estimation. LLMs are used without RAG for replays. ML receives weld classes only, while LLMs also receive quantitative feedback where available.}
\label{tab:cross-evaluation}
\footnotesize
\setlength{\tabcolsep}{3pt}
\renewcommand{\arraystretch}{1.15}
\begin{tabularx}{\columnwidth}{@{}>{\raggedright\arraybackslash}p{0.17\columnwidth} Y Y@{}}
\toprule
& LLM lobe estimation & ML lobe estimation \\
\midrule
LLM trial selection
& \textbf{Closed loop}\newline
Gemma, Claude, GPT
& \textbf{Replay}\newline
Gemma $\rightarrow$ Bayes / Deep Sets\newline
Claude $\rightarrow$ Bayes / Deep Sets\newline
GPT $\rightarrow$ Bayes / Deep Sets \\
\addlinespace
ML trial selection
& \textbf{Replay}\newline
IG $\rightarrow$ Gemma / Claude / GPT\newline
PPO $\rightarrow$ Gemma / Claude / GPT
& \textbf{Closed loop}\newline
IG $\rightarrow$ Bayes\newline
PPO $\rightarrow$ Deep Sets \\
\bottomrule
\end{tabularx}
\end{table}

Cross-evaluation is used to examine LLM trial selection and lobe estimation separately (Table~\ref{tab:cross-evaluation}). To assess trial selection, we provide LLM-generated trial records to the Bayesian and Deep Sets estimators and compare their lobe predictions with those obtained from ML-generated records using the same estimator. To assess lobe estimation, we provide ML-generated records to Gemma, Claude, and GPT and compare their predictions with the ML estimates from the same records. These comparisons use LLMs without RAG. LLMs receive weld classes and available quantitative responses, whereas ML methods receive weld classes only.

\subsection{Lobe transfer across electrode-force conditions}
\label{sec:transfer-method}

In addition to exploring lobes through trial welds, we evaluate whether LLMs can use previously estimated lobes to predict a lobe under a different electrode force without trial welds at that force. Each LLM is provided with two lobes previously estimated by the same model and configuration, their corresponding electrode forces, and a new electrode force, and is asked to predict the lobe under the new force. We use the seven electrode-force conditions in the Shin dataset to assess this transfer capability and evaluate the predicted lobes using mean interval IoU.

\subsection{Welding schedule generation, outcome prediction, and revision}
\label{sec:initial-method}

Three further tasks test the LLMs without the closed loop: welding schedule generation, outcome prediction, and revision after a failed trial.

For welding schedule generation, the LLM receives the same case description and acceptance requirement as in Section~\ref{sec:method}. Each model and configuration proposes a welding current and welding time without any trial results, in three independent fresh chats per lobe, giving 33 proposed schedules per model and configuration. A proposed schedule is considered successful if its current--time setting lies within the lobe.

We further assess the ability of LLMs to predict how weld outcomes change with current at a specified welding time. The same case description and acceptance requirement as in schedule generation are provided, and the welding time is additionally specified. This task is evaluated on the Shin dataset across seven electrode forces and three welding times (300, 400, and 500~ms), yielding 21 prediction tasks per model and configuration. We report weld outcome classification accuracy and the mean absolute error (MAE) of tensile-shear load predictions. For the current intervals derived from these classifications, we calculate interval IoU against the reference intervals in Fig.~\ref{fig:outcome-windows} and average across the seven electrode forces and three welding times.

For schedule revision, we provide the trial schedule, weld class (undersized, acceptable, or expulsion), and, where available, the estimated quantitative weld response (tensile-shear load or nugget diameter). For each lobe, we select failed starting points at five welding times, covering four positions at each time: near and far below the lower boundary, and near and far above the upper boundary. For each starting point, we conduct three independent fresh chats and request a revised current--time setting, allowing changes to the current, welding time, or both. Revision success is determined using the same acceptance criterion as schedule generation. Revision success rates and the direction and magnitude of changes in welding current and time are reported.

\subsection{Weld classification and response prediction from images}
\label{sec:image-method}

The virtual laboratory directly provides weld classes and quantitative responses. However, using collected weld images as feedback in autonomous experiments entails visual interpretation. We therefore evaluate LLM performance in weld classification and quantitative response estimation from trial-weld photographs. We use the Resistance Spot Welding Insights dataset~\citep{dominguez2025data}, which provides paired surface photographs and measured outcomes for 495 AISI 1010 steel welds. For each weld, we provide front and back photographs and request its class, nugget diameter, and pull force. Predictions without examples are compared with predictions using 24 labeled welds, eight per class, and both are evaluated on the remaining 471 welds in fresh chats. These tests use no RAG and are evaluated separately from lobe exploration. We report classification accuracy, per-class recall, and MAE for nugget diameter and pull force. For comparison, we train a convolutional neural network (CNN) from scratch, using either the same 24 examples or five-fold cross-validation on the full dataset, and score it on the same 471 welds.

\section{Results}
\label{sec:results}

{\widowpenalty=10000
This section first presents the lobe exploration results and the comparison with ML methods, followed by lobe transfer across electrode-force conditions.  It then reports schedule generation, outcome prediction, and revision, which show what the LLMs can do without the closed loop, and weld classification and quantitative response prediction from images.\par}

\begin{figure}[!t]
  \centering
  \includegraphics[width=\columnwidth]{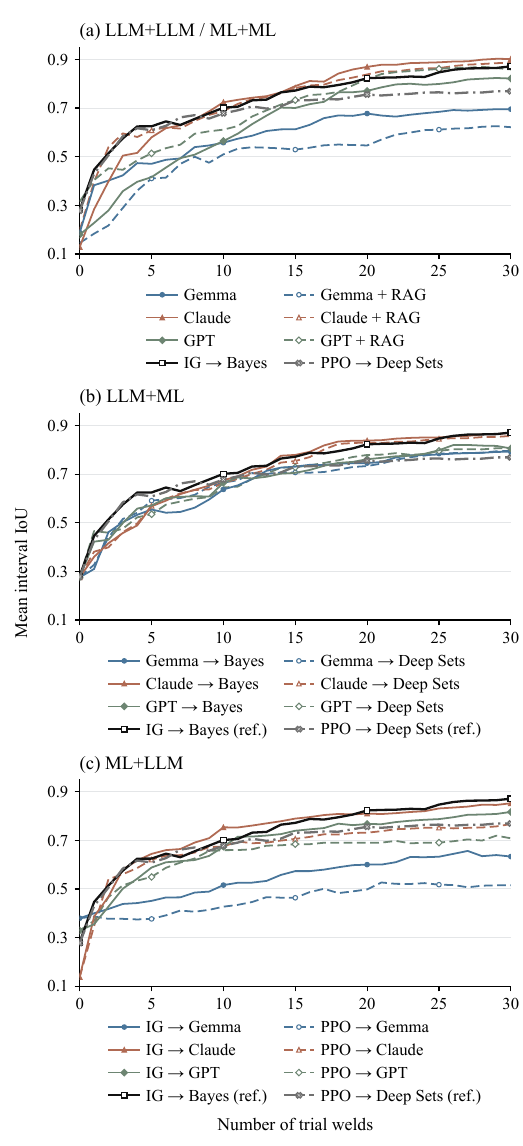}
  \caption{Lobe exploration over eleven reference lobes. (a) LLM and ML closed loops. (b) LLM-selected records evaluated by ML estimators. (c) ML-selected records evaluated by LLMs. Legend arrows denote trial selection followed by lobe estimation. IG denotes Bayesian information-gain selection. ML reference curves use run~1 per lobe, whose records are replayed in (c), and are repeated across panels. Panels (b,c) use LLMs without RAG and fixed-record replay. LLMs receive quantitative feedback where available; ML methods receive classes only. Curves average mean interval IoU over the eleven lobes, using their original measured welding times.}
  \label{fig:llm-discovery}
  \label{fig:closed-loops}
  \label{fig:selection-estimation}
\end{figure}

\subsection{Lobe exploration performance and comparison with ML methods}
\label{sec:sequential-results}

Lobe predictions improve as trial results accumulate, as shown in Fig.~\ref{fig:llm-discovery}a. Without RAG, mean IoU increases from 0.188 at zero trials to 0.695 after 30 trials for Gemma, from 0.128 to 0.902 for Claude, and from 0.179 to 0.821 for GPT. RAG yields higher IoUs for Claude and GPT during the first five trials, whereas Gemma shows no such early benefit. After 30 trials, IoUs with RAG are 0.621, 0.886, and 0.859 for Gemma, Claude, and GPT, respectively.  RAG therefore lowers the final IoU for Gemma and Claude and raises it for GPT, with no consistent benefit.

The ML closed loop using information gain (IG) for trial selection and Bayesian estimation for lobe estimation reaches a mean IoU of 0.871 after 30 trials. The ML closed loop using proximal policy optimization (PPO) for trial selection and a Deep Sets estimator for lobe estimation reaches 0.769. At the same trial budget, Claude without RAG achieves 0.902, which is 0.031 and 0.133 above these two ML baselines, respectively. Gemma without RAG reaches 0.695, remaining below both baselines. GPT without RAG reaches 0.821, placing it between the two ML baselines.  In these comparisons, the LLMs also receive quantitative weld responses where available, whereas the ML methods receive weld classes only.

To examine trial selection and lobe estimation separately, we cross-evaluate the LLM and ML methods using recorded trial histories. For trial selection, we compare records selected by LLMs without RAG with those selected by ML methods, using the same ML lobe estimator, as shown in Fig.~\ref{fig:selection-estimation}b. With Bayesian estimation, records selected by Gemma, Claude, and GPT yield final IoUs of 0.792, 0.870, and 0.806, respectively, compared with 0.871 for IG-selected records. Claude-selected records therefore support an estimate close to the IG baseline, whereas Gemma- and GPT-selected records yield lower IoUs. With Deep Sets estimation, the same LLM-selected records yield final IoUs of 0.796, 0.857, and 0.807, respectively, all exceeding the 0.769 obtained from PPO-selected records. Claude-selected records yield the highest final IoU among the three LLM record sources with both ML estimators.

To evaluate lobe estimation, we compare LLM and ML estimates obtained from the same ML-selected trial records, as shown in Fig.~\ref{fig:selection-estimation}c. On IG-selected records, Gemma, Claude, and GPT without RAG achieve final IoUs of 0.633, 0.852, and 0.817, respectively, compared with 0.871 for Bayesian estimation. On PPO-selected records, the corresponding values are 0.515, 0.767, and 0.708, compared with 0.769 for Deep Sets. All three LLMs obtain lower final IoUs than the corresponding ML estimator on both record sets. Claude comes closest to the ML estimators, with differences of 0.019 on IG-selected records and 0.002 on PPO-selected records, followed by GPT and Gemma. Together, these results show that Claude can both select informative trial schedules and estimate lobes from trial records at a level close to the ML baselines, while GPT and Gemma trail the ML estimators by larger margins.

\subsection{Lobe transfer performance across electrode-force conditions}
\label{sec:transfer-results}

\begin{figure}[!tbp]
  \centering
  \includegraphics[width=\columnwidth]{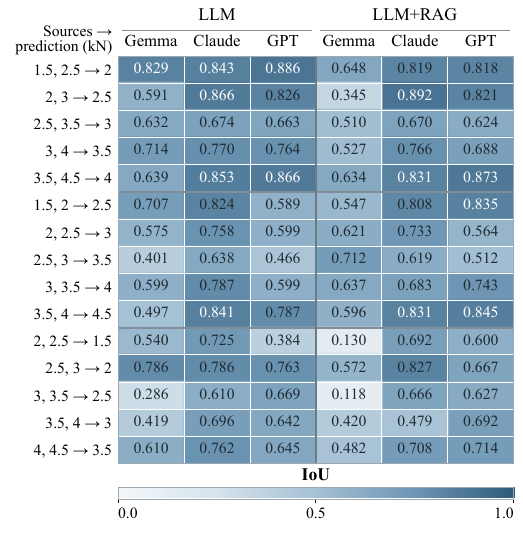}
  \caption{Lobe transfer across electrode-force conditions. Each source--target label gives the two source forces followed by the target force, in kN. The fifteen combinations include five with source forces bracketing the target, five with both source forces below the target, and five with both above. Each cell reports mean interval IoU at the original measured welding times. Source lobes are obtained through lobe exploration using the same model and configuration.}
  \label{fig:transfer-direction}
\end{figure}

We evaluate the ability of LLMs to transfer lobe estimates across electrode-force conditions. Fifteen transfer tasks are constructed from the Shin aluminum--steel dataset~\citep{shin2019}. In each task, the model receives two previously estimated source lobes and predicts a lobe at a new electrode force without trial welds at that force. Without RAG, Gemma, Claude, and GPT achieve mean IoUs of 0.588, 0.762, and 0.677, respectively, as shown in Fig.~\ref{fig:transfer-direction}. With RAG, the corresponding scores are 0.500, 0.735, and 0.708. Claude achieves the highest overall IoU in both configurations.

Without RAG, all three models perform best when the target force lies between the two source forces, achieving mean IoUs of 0.681, 0.801, and 0.801 for Gemma, Claude, and GPT, respectively. When both source forces are below the target, their scores are 0.555, 0.770, and 0.608. When both are above, the scores are 0.528, 0.716, and 0.621. Claude and GPT perform similarly for targets between the source forces, while Claude achieves higher IoUs for targets outside the source-force range. RAG improves lobe-transfer performance for GPT but reduces it for Gemma and Claude. Because each configuration uses its own previously estimated source lobes, these scores reflect both the quality of the source lobes and the transfer step itself.

Together, the exploration and transfer results indicate that LLMs can support trial selection, lobe estimation, and the reuse of exploration results under changed operating conditions.

\subsection{Performance in schedule generation, outcome prediction, and revision}
\label{sec:initial-results}

\begin{figure*}[!tp]
  \centering
  \includegraphics[width=\textwidth]{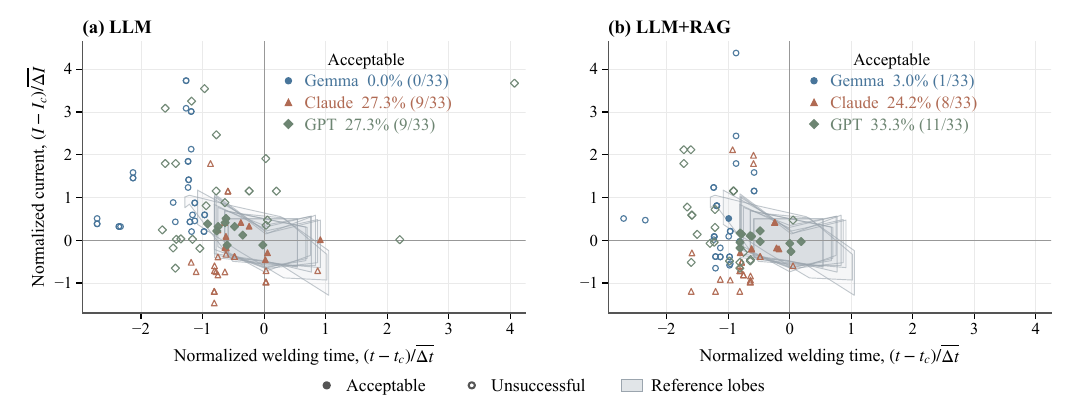}
  \caption{Welding schedule generation: (a) LLM and (b) LLM+RAG. Each case is centered at its reference-lobe centroid and normalized by its mean time extent and current range. Gray polygons show the eleven reference lobes. Filled markers indicate schedules acceptable for their corresponding cases and hollow markers indicate unsuccessful schedules.}
  \label{fig:initial-selection}
\end{figure*}

In these tasks, the LLMs receive no trial results at the target condition, or only one failed trial.  Schedule generation with and without RAG has low success rates across the eleven reference lobes, as shown in Fig.~\ref{fig:initial-selection}. Success rates for Gemma, Claude, and GPT are 0\%, 27.3\%, and 27.3\% without RAG, and 3.0\%, 24.2\%, and 33.3\% with RAG, respectively. Most recommendations fall outside the reference lobes, and RAG does not consistently improve the success rate. Some recommendations are far from the lobe.  Without RAG, GPT recommends 900~ms at 16~kA and 650~ms at 8.5~kA for the aluminum--steel cases at electrode forces of 1.5 and 2.0~kN, respectively, with welding times well beyond the tested range of 300--500~ms.

Each model shows a characteristic bias.  Gemma favors short welding times and relatively high currents.  GPT selects welding times within the reference ranges for most lobes, but without RAG its currents tend to be high.  Claude selects similar welding times with and without RAG but often chooses currents below the lower lobe boundary. 

For outcome prediction on the aluminum--steel dataset reported by \citet{shin2019}, classification accuracies without RAG are 41.9\%, 61.4\%, and 57.1\% for Gemma, Claude, and GPT, respectively. For quantitative weld-response prediction, tensile-shear load MAEs are 1142, 600, and 922~N, respectively, against an acceptance threshold of 3000~N. With RAG, classification accuracies become 49.5\%, 61.9\%, and 56.2\%, while load MAEs increase to 1428, 681, and 963~N. Claude achieves the highest classification accuracy and lowest load MAE in both configurations. RAG improves classification for Gemma, changes it little for Claude and GPT, and increases load prediction errors for all three models. Access to the retrieved literature therefore does not make outcome prediction accurate for a given schedule.

\begin{figure*}[!tp]
  \centering
  \includegraphics[width=\textwidth]{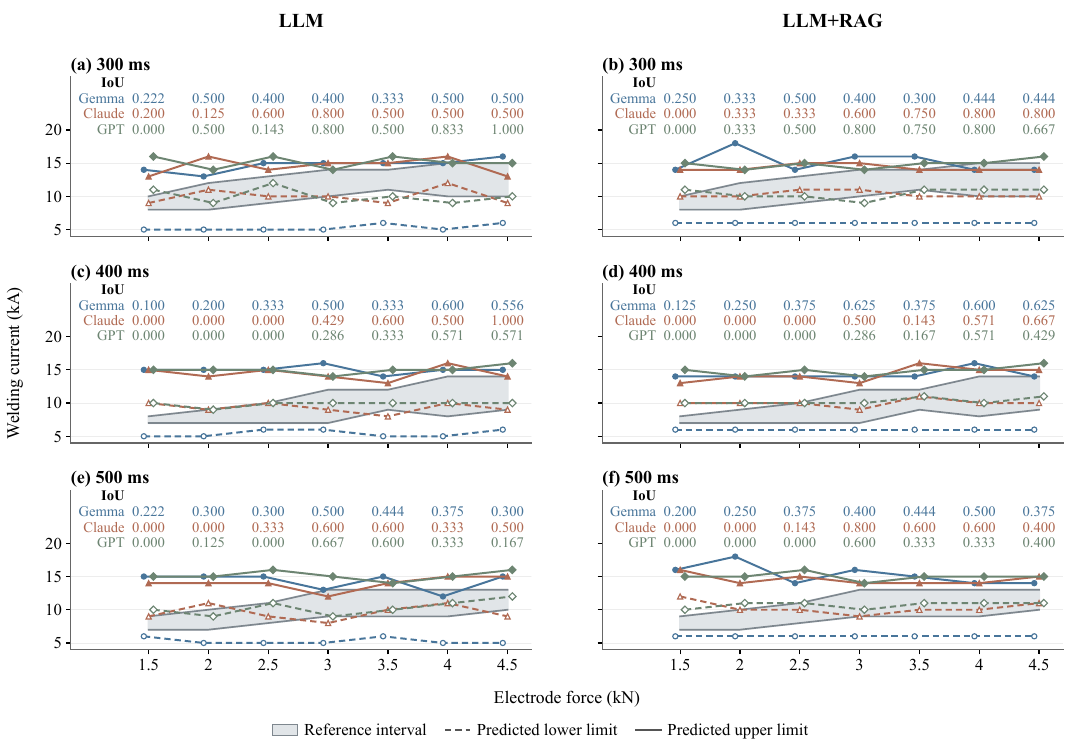}
  \caption{Acceptable-current intervals derived from outcome predictions: (a,b) 300~ms, (c,d) 400~ms, and (e,f) 500~ms. Panels (a,c,e) show LLM predictions, and (b,d,f) show LLM+RAG predictions. Gray bands show reference intervals across electrode forces. In (a,b), missing upper reference limits at 4.0 and 4.5~kN are completed at 15~kA using the virtual-laboratory reconstruction. Dashed and solid colored lines indicate predicted lower and upper limits, respectively. Model curves are slightly offset horizontally for visibility. In-panel values report interval IoU for each model and electrode force, computed from the predicted and displayed reference intervals.}
  \label{fig:outcome-windows}
\end{figure*}

As shown in Fig.~\ref{fig:outcome-windows}, the acceptable-current intervals predicted for the Shin aluminum--steel dataset~\citep{shin2019} show limited overlap with the reference intervals across the three welding times and three LLMs. Averaged over the seven electrode forces and three welding times, mean interval IoUs without RAG are 0.377, 0.387, and 0.354 for Gemma, Claude, and GPT, respectively. With RAG, the corresponding values are 0.390, 0.383, and 0.332. Both with and without RAG, the predictions reflect the basic progression from undersized welds to acceptable welds and then expulsion as welding current increases. However, the predictions do not consistently reproduce how the reference intervals change with welding time. The predictions also show limited sensitivity to changes in electrode force.

\begin{figure*}[!tp]
  \centering
  \includegraphics[width=\textwidth]{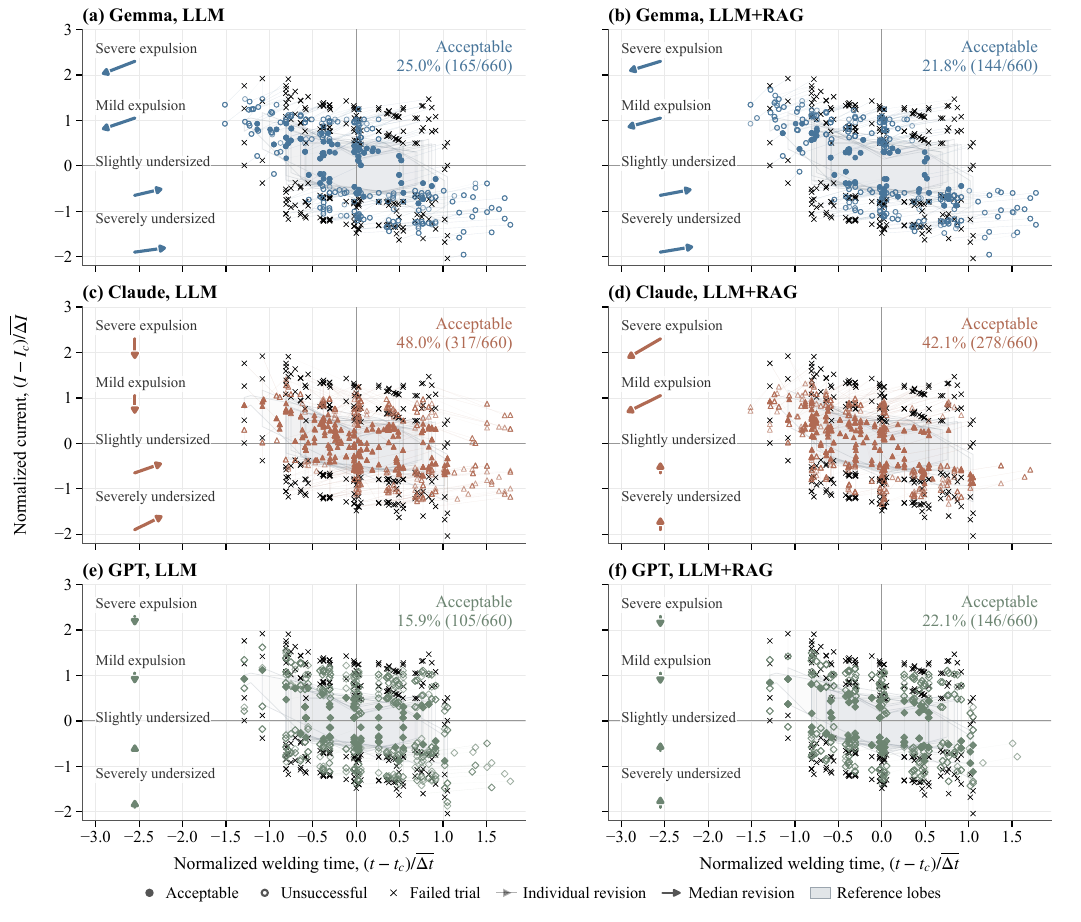}
  \caption{Schedule revision after one failed trial: (a,b) Gemma, (c,d) Claude, and (e,f) GPT. Panels (a,c,e) show LLM results, and (b,d,f) show LLM+RAG results. Gray polygons show the eleven reference lobes. Black crosses mark failed trials, and faint arrows show individual revisions. Colored markers show revised schedules, filled when acceptable and hollow otherwise. Each bold arrow shows the coordinate-wise median normalized time and current changes for one starting-condition group of 165 revisions. Mild/slight and severe denote near- and far-boundary starts, respectively. Summary arrows are translated to the left for visibility, preserving their direction and magnitude. Percentages and counts report successful revisions out of 660 responses per model and configuration. Repeated coordinates overlap.}
  \label{fig:feedback-revision}
\end{figure*}

Revisions of failed trial schedules are evaluated for undersized welds and expulsion, each at mild and severe levels. As shown in Fig.~\ref{fig:feedback-revision}, revision success rates without RAG are 25.0\% for Gemma, 48.0\% for Claude, and 15.9\% for GPT. With RAG, these rates become 21.8\%, 42.1\%, and 22.1\%, respectively. Success is lower for severe failures, whose starting schedules deviate more from their reference lobes.  For example, for Claude without RAG, success decreases from 61.8\% for mildly undersized starts to 33.3\% for severely undersized starts, and from 80.0\% for mild expulsion to 17.0\% for severe expulsion.

The median revision vectors in Fig.~\ref{fig:feedback-revision} reveal distinct adjustment tendencies among the models. Gemma generally changes current and welding time together, increasing both after undersized welds and decreasing both after expulsion. Its welding-time adjustments are similar for mild and severe failures. Claude generally makes larger current corrections, with greater increases for severely undersized starts, and combines these with welding-time changes in some cases. GPT is more conservative, predominantly adjusting current while retaining the original welding time. Although the models generally adjust current in the expected direction, the corrections frequently remain insufficient to reach the acceptable region, particularly for severe failures.

\begin{figure*}[!t]
  \centering
  \includegraphics[width=\textwidth]{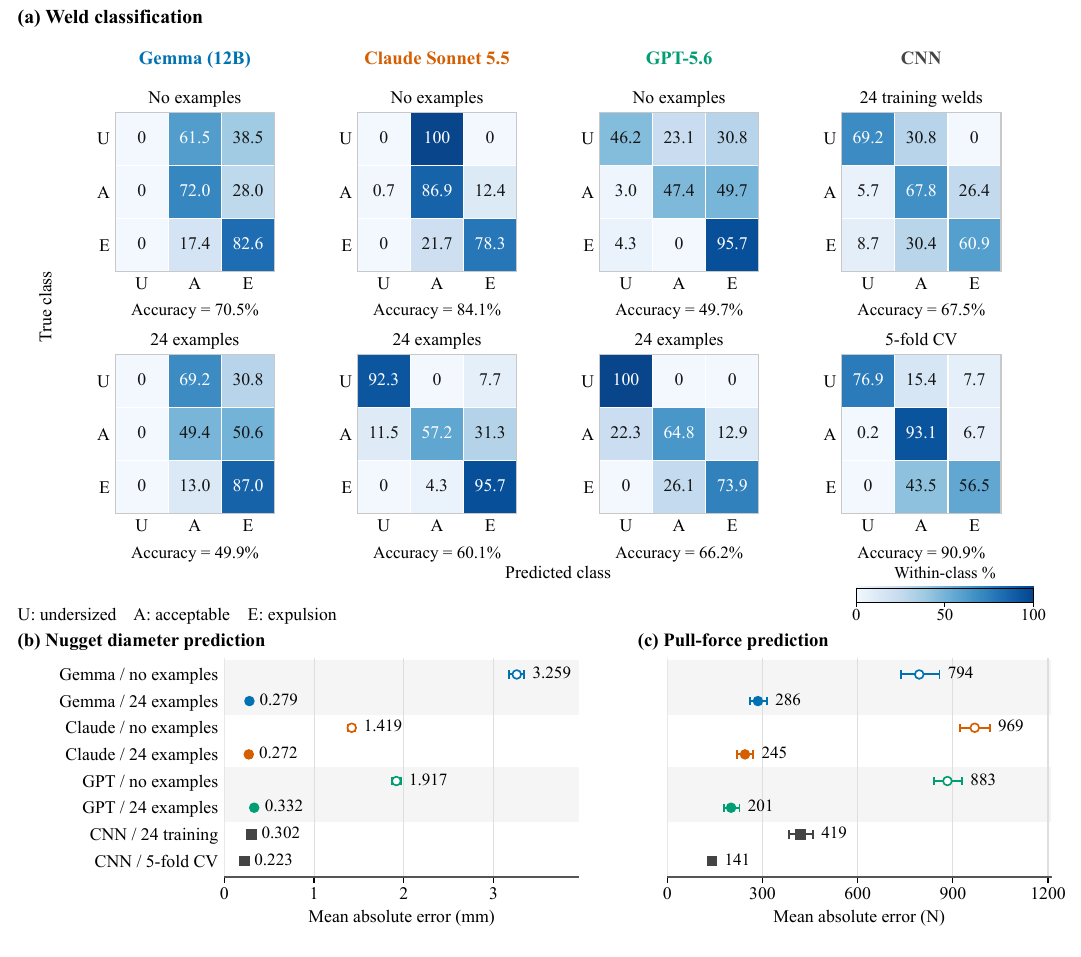}
  \caption{Weld classification and response prediction from images on 471 welds: 13 undersized, 435 acceptable, and 23 expulsion. (a) Confusion matrices normalized within each true class, with overall accuracy below each matrix. Rows indicate true classes and columns indicate predictions. LLMs receive no examples or 24 labeled examples. CNN configurations use 24 training welds or more training data under five-fold cross-validation. (b,c) Nugget-diameter and pull-force MAEs with 95\% bootstrap intervals.}
  \label{fig:weld-inspection}
\end{figure*}

\subsection{Performance in weld classification and response prediction from images}
\label{sec:inspection-results}

Figure~\ref{fig:weld-inspection} shows image-based classification results for 471 welds from the Resistance Spot Welding Insights dataset~\citep{dominguez2025data}. Without examples, classification accuracies are 70.5\%, 84.1\%, and 49.7\% for Gemma, Claude, and GPT, respectively.  Because acceptable welds account for 435 of the 471 welds, overall accuracy mainly reflects acceptable-weld recall, and the per-class recall in Fig.~\ref{fig:weld-inspection}a gives a fuller comparison.  Despite achieving the highest overall accuracy, Claude misclassifies all 13 undersized welds as acceptable. With 24 labeled examples, classification accuracies become 49.9\%, 60.1\%, and 66.2\%, respectively. Undersized-weld recall increases from 0 to 92.3\% for Claude and from 46.2\% to 100\% for GPT, while Gemma detects no undersized welds in either configuration. However, acceptable-weld recall decreases from 72.0\% to 49.4\% for Gemma and from 86.9\% to 57.2\% for Claude. Gemma classifies more acceptable welds as expulsion, while Claude classifies more as either undersized or expulsion. These additional errors on the dominant class reduce overall accuracy for both models. GPT instead improves acceptable-weld recall from 47.4\% to 64.8\%, primarily by classifying fewer acceptable welds as expulsion.

As shown in Fig.~\ref{fig:weld-inspection}b,c, providing 24 labeled examples reduces quantitative prediction errors for all three LLMs. Nugget-diameter MAE falls from 1.419--3.259~mm without examples to 0.272--0.332~mm with examples, and pull-force MAE falls from 794--969~N to 201--286~N.

Two CNN configurations are evaluated: one trained on the same 24 labeled welds provided as examples to the LLMs, and another trained on four of the five folds of the full dataset under five-fold cross-validation. With the same 24 labeled welds, the CNN achieves 67.5\% classification accuracy, exceeding all three LLMs with examples in overall accuracy, although Claude and GPT achieve higher undersized- and expulsion-weld recall. However, Gemma and Claude achieve lower nugget-diameter MAEs of 0.279 and 0.272~mm, respectively, compared with 0.302~mm for the CNN. All three LLMs also achieve lower pull-force MAEs than the CNN value of 419~N, with GPT achieving the lowest error of 201~N among the configurations using 24 labeled welds. Under five-fold cross-validation with more training data, the CNN achieves 90.9\% classification accuracy, a nugget-diameter MAE of 0.223~mm, and a pull-force MAE of 141~N, outperforming all evaluated LLM configurations on these three metrics, although Claude and GPT with 24 examples detect undersized and expulsion welds more often.

\section{Discussion}
\label{sec:discussion}

The results show that closed-loop exploration is where LLMs are useful in RSW process development.  Direct schedule recommendations and weld-response predictions remain unreliable, and retrieved literature does not consistently improve them, but with trial feedback Claude and GPT improve their lobe estimates and choose informative next trials.  The cross-evaluation shows that the final lobe depends on both which trials are run and how their results are interpreted.  Trial records selected by Gemma yield IoUs of 0.792 and 0.796 with the Bayesian and Deep Sets estimators, compared with 0.695 from Gemma's own estimates, so Gemma does not fully use the information in its own trials, whereas Claude performs well in both roles.  Evaluating the two roles separately shows where an exploration system needs improvement and supports hybrid designs that pair one method's trial selection with another's lobe estimation.

General-purpose LLMs also need little preparation for a new task.  The ML methods in this study depend on predefined lobe representations, prior assumptions, or synthetic training distributions, which may need adjustment when operating conditions or lobe shapes change substantially.  An LLM receives the task, operating conditions, and trial records as text and can take on several related decisions without additional training.  Its practical value should therefore be judged by prediction error and trial efficiency together with the modeling and training effort that a new task requires.  The transfer results show that a lobe obtained through exploration can inform predictions under changed operating conditions.  Errors in the estimated lobes carry over, however, so the reliability of transfer is tied to the quality of the preceding exploration.

This study has several limitations.  The decision tasks are not yet connected into a complete closed loop in physical welding.  Exploration uses structured feedback from the virtual laboratory, while photograph interpretation is evaluated separately on a different dataset, so the results do not show how observation errors would affect later trial selection or accumulate over successive lobe updates.  The virtual laboratory's feedback and scoring both rely on reconstructed process responses, and it does not include experimental scatter, electrode wear, or process drift.  Boundary identification in physical trials may therefore require repeated measurements and explicit treatment of uncertainty.  The transfer experiments examine only electrode-force changes within the same material and equipment system.  The image evaluation is limited by its specific acquisition conditions and the small number of defective welds.  Finally, the reference lobes are taken from published open-access studies that may appear in the models' training data.  The low IoUs before any trials (0.128--0.188 without RAG) suggest that the models do not recall these lobes directly, but this possibility cannot be ruled out.

Future work should connect physical welding trials, image acquisition, quality measurements, and LLM decisions within a single experimental system and track how observation errors and process variability affect trial selection, lobe boundaries, and stopping decisions.  Exploration procedures reviewed by welding engineers should also be evaluated through repeated runs under a fixed trial budget.  Experiments that vary materials, sheet thicknesses, equipment, and pulse schedules would show how far the same LLM workflow can be reused and what additional process information it requires.

\section{Conclusion}
\label{sec:conclusion}

This paper proposes a closed-loop weld lobe development method for RSW using LLMs.  In each iteration, the LLM selects the welding schedule for the next trial weld and updates its lobe prediction from the trial results collected so far, without task-specific training.  The method is evaluated with Gemma, Claude Sonnet 5.5, and GPT-5.6 on eleven reference lobes, using a virtual trial-welding laboratory built from published data.
After 30 trial welds without RAG, the mean interval IoUs are 0.695, 0.902, and 0.821 for Gemma, Claude, and GPT, respectively, compared with 0.871 and 0.769 for the two ML closed loops.  Cross-evaluation separates trial selection from lobe estimation and shows that Claude performs close to the ML baselines in both roles.  Lobes obtained through exploration can also support predictions under a changed electrode force.  Without trial feedback, schedule generation and outcome prediction remain limited, and RAG does not consistently improve performance.  In a separate image-based test, labeled examples reduce nugget-diameter and pull-force prediction errors.

These results show that general-purpose LLMs can develop weld lobes from trial feedback about as accurately as specialized ML methods.  The same approach could be used for other process development tasks in manufacturing where a process window is found through sequential experiments.

\section*{Acknowledgments}

The authors acknowledge the financial support of General Motors (GM).  Z. Yang gratefully acknowledges the support of the Chair's Fellowship from the Department of Mechanical Engineering at the University of Michigan.

\balance
\bibliographystyle{elsarticle-num-names}
\bibliography{references}

\end{document}